\pdfoutput=1

\documentclass[11pt]{article}

\usepackage[final]{acl}

\usepackage{times}
\usepackage{amsmath}
\usepackage{latexsym}
\usepackage{booktabs}
\usepackage{hyperref} 
\usepackage{tcolorbox}
\usepackage{subcaption}
\usepackage{array}
\usepackage{xcolor}
\usepackage{soul}
\usepackage{multicol,multirow}
\usepackage{todonotes}
\usepackage{xspace}
\definecolor{mygold}{HTML}{eeba0a}
\definecolor{mygrey}{HTML}{bac8ca}
\definecolor{mylightgreen}{HTML}{D1F694}
\definecolor{myblue}{HTML}{30C0F0}
\definecolor{myorange}{HTML}{F5AF22}
\definecolor{myred}{HTML}{FF576A}
\definecolor{darkred}{HTML}{91270F}
\definecolor{darkgreen}{HTML}{5E893E}
\definecolor{lightblue}{HTML}{b6ecfc}
\sethlcolor{mylightgreen}
\usepackage{makecell}
\usepackage{longtable}
\usepackage{amssymb}
\usepackage{pifont}
\definecolor{lightpink}{rgb}{1.0, 0.8, 0.9}

\usepackage[T1]{fontenc}

\usepackage[utf8]{inputenc}

\usepackage{microtype}

\usepackage{inconsolata}

\usepackage{graphicx}
\usepackage{algorithm}
\usepackage{algpseudocode}

\title{NewsRECON: News article REtrieval for image CONtextualization}

\author{
Jonathan Tonglet$^{1,2,3}$, Iryna Gurevych$^{1}$,  Tinne Tuytelaars$^{2}$, Marie-Francine Moens$^{3}$ 
\\
        \textsuperscript{1}Ubiquitous Knowledge Processing Lab (UKP Lab), Department of Computer Science, \\ TU Darmstadt and National Research Center for Applied Cybersecurity ATHENE\\ 
\textsuperscript{2}Department of Electrical Engineering, KU Leuven\\
\textsuperscript{3}Department of Computer Science, KU Leuven\\
\href{jonathan.tonglet@kuleuven.be}{jonathan.tonglet@kuleuven.be}
}

\begin{document}
\maketitle
\begin{abstract}
Identifying when and where a news image was taken is crucial for journalists and forensic experts to produce credible stories and debunk misinformation. While many existing methods rely on reverse image search (RIS) engines, these tools often fail to return results, thereby limiting their practical applicability. In this work, we address the challenging scenario where RIS evidence is unavailable. We investigate the potential of news article corpora as an alternative to RIS, linking images to relevant articles to infer their dates and locations from article metadata. We evaluate the performance of a news article retrieval pipeline, NewsRECON, which leverages a corpus of over 85,000 articles. Experiments on the TARA dataset show that NewsRECON outperforms prior work and can be combined with a multimodal large language model (MLLM) to achieve new SOTA results in the absence of RIS evidence. Furthermore, NewsRECON generalizes to the
5Pils-OOC benchmark despite geographic and temporal shifts. We make our code available.\footnote{\href{https://github.com/jtonglet/emnlp2026-newsrecon}{github.com/jtonglet/emnlp2026-newsrecon}}

\end{abstract}

\section{Introduction}

Images are frequently used to communicate news events. However, images are often shared with missing or incorrect metadata, such as their date or location. These inaccuracies play a significant role in the spread of misinformation in the news domain \citep{dufour2024ammeba,cao2025averimatec,geng2025m4fc}. As a result, journalists, fact-checkers, and forensic experts must frequently verify or recover the metadata of images, i.e., \textit{image contextualization}, a task that is crucial for reliable news coverage, investigation of war crimes \citep{silverman2013verification}, or early misinformation detection \citep{puliafito2025,khan2024debunking}. A common approach involves using RIS tools to retrieve web pages that contain the image and to infer the date or location from the page content. However, RIS engines often fail to return results \citep{tonglet-etal-2024-image}. In such cases, identifying the date and location becomes more difficult, requiring journalists to interpret visual cues and gather external evidence. Due to time and resource constraints, these cases are also more interesting to automate.

\begin{figure}
    \centering
    \includegraphics[width=\linewidth]{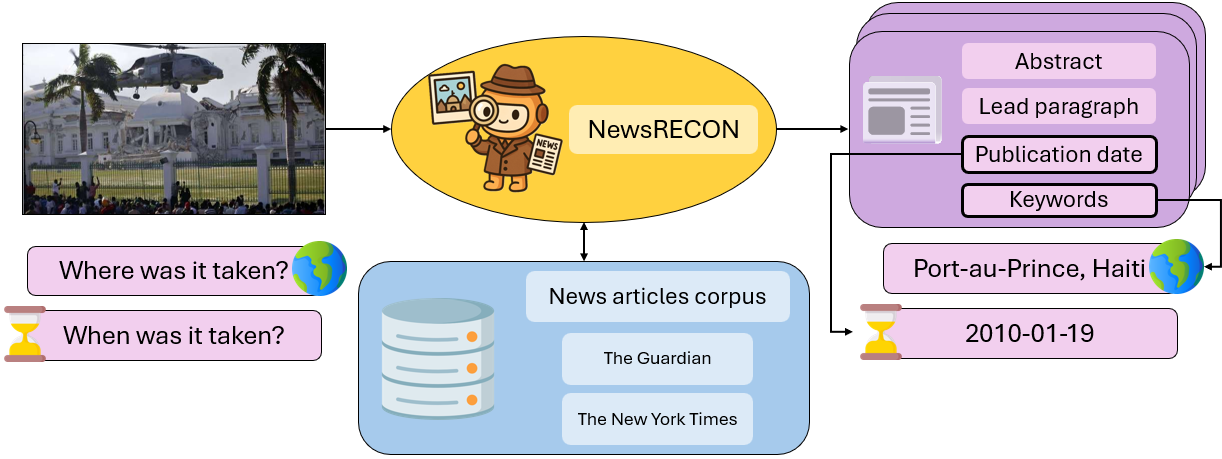}
    \caption{The metadata of location and event-relevant news articles can be used to contextualize a news image.}
    \label{fig:task}
\end{figure}

Several works have proposed methods to automate image contextualization \citep{fu-etal-2022-theres,tonglet-etal-2024-image,10.1145/3746027.3767108}. Some of them address the challenging scenario in which RIS engines are unavailable, relying solely on the image \citep{fu-etal-2022-theres,siingh-etal-2025-getreason} or on encyclopedic knowledge bases \citep{10.1145/3689638}. However, none of these methods leverage large-scale news article corpora, which are a rich source of contextual evidence \citep{fu-etal-2022-theres}.

In this work, we investigate the potential of leveraging news article corpora for image contextualization in settings where RIS engines are unavailable. To this end, we evaluate a cross-modal retrieval pipeline, NewsRECON, illustrated in Figure \ref{fig:task}. NewsRECON searches a large corpus to identify relevant articles for a given image, i.e., articles relating to the same location or event, where an event is defined as the combination of a date and a location. The metadata of these articles can then be used as predictions of the image's date and location. We construct a corpus of 88,859 news articles spanning the years 2010 to 2023. The evaluated pipeline comprises three main components: (1) a bi-encoder that retrieves the top-K event-relevant articles, (2) a cross-encoder that reranks these articles based on their location similarity to the image, and (3) a second cross-encoder that reranks clusters of articles by their event consistency with the image. Experiments on the TARA benchmark show that this approach outperforms retrieval baselines by up to 11 percentage points (pp) in GREAT score \citep{siingh-etal-2025-getreason}, a joint metric for date and location prediction. Moreover, providing the top-3 retrieved articles as context to a MLLM yields new state-of-the-art results. We further demonstrate that NewsRECON, trained on TARA, generalizes to 5Pils-OOC \citep{tonglet-etal-2025-cove}, despite geographic, temporal, and stylistic shifts. When combined with an MLLM, it achieves the strongest results for images lacking RIS web evidence. In summary, this work presents the first investigation of news article corpora as external evidence for image contextualization, demonstrating their value as a viable alternative to RIS engines.

\section{Related work}

Journalists and forensic experts contextualize news images by identifying key context items such as the source, date, location, and motivation \citep{urbani2020verifying,tonglet-etal-2024-image}. Applications of this process include fact-checking images taken out of context to spread misinformation \citep{tonglet-etal-2025-cove} and identifying evidence for war crimes investigations \citep{silverman2013verification}. In this work, we focus on two context items: the date and location, which are the most widely studied.

Several datasets have been proposed for this task \citep{fu-etal-2022-theres,tonglet-etal-2024-image,geng2025m4fc}.
Existing methods differ in both their model architectures and the types of external evidence they use. Some approaches formulate image contextualization as a classification task, fine-tuning CLIP \citep{pmlr-v139-radford21a} to retrieve the best matching candidate date-location pairs within the same dataset split \citep{fu-etal-2022-theres,10.1145/3689638}. However, this setup lacks realism: the candidate set is limited to a small number of date-location pairs from the split, whereas in real-world scenarios, there are thousands of unique dates and an even larger number of locations, resulting in a near-infinite search space.

Other works treat it as a text generation task and rely on (M)LLMs to generate the answers \citep{tonglet-etal-2024-image,10.1145/3746027.3767108,10484512,ayyubi-etal-2025-puzzlegpt,siingh-etal-2025-getreason,tonglet-etal-2025-cove}.  Some of these approaches use RIS engines to retrieve external evidence \citep{tonglet-etal-2024-image,10.1145/3746027.3767108,tonglet-etal-2025-cove}. However, RIS engines may return no results \citep{Abdelnabi_2022_CVPR,tonglet-etal-2024-image}. The absence of RIS evidence makes the task far more challenging but also more valuable to automate, as it requires interpreting subtle visual cues and relating them to world knowledge. Other methods explore external evidence, such as Wikipedia \citep{10.1145/3689638,tonglet-etal-2025-cove} or celebrity recognition APIs \citep{siingh-etal-2025-getreason}. However, none of the existing methods leverage news articles, despite their strong potential as a source of temporal and spatial context \citep{fu-etal-2022-theres}.

\section{NewsRECON}

\begin{figure*}
    \centering
    \includegraphics[width=0.85\linewidth]{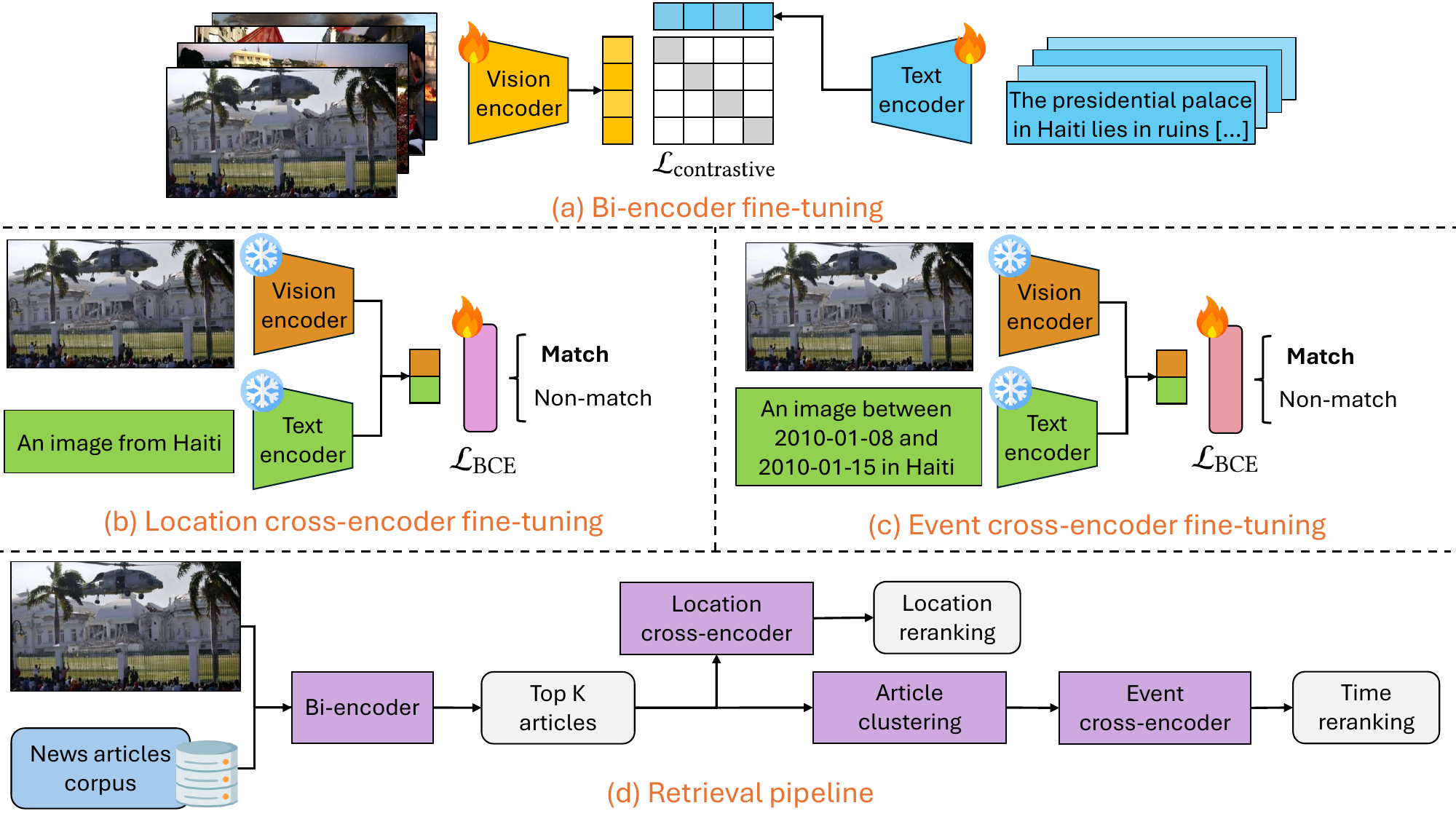}
    \caption{Overview of NewsRECON. Parts a to c illustrate the three training stages with images from TARA \citep{fu-etal-2022-theres}. Part d shows the retrieval pipeline at inference time.}
    \label{fig:newsrecon}
\end{figure*}

We use a cross-modal retrieval pipeline, NewsRECON, illustrated in Figure~\ref{fig:newsrecon}, to investigate the potential of news articles as external evidence for image contextualization. Given an image, it returns a ranked list of relevant news articles, whose metadata can be used to predict the image's date and location. The inference-time pseudocode is provided in Appendix~\ref{sec:pseudocode}.

\paragraph{Relevant article labeling}
For each image, we define two sets of relevant articles that serve as weak training supervision signals: one for location and one for event (date + location), with the event set being a subset of the location set. An article is location-relevant if at least one of its geolocation keywords contains the ground truth location. Event relevance also requires that the article's publication date fall within a window of ±$N_\text{window}$ days around the image's publication date, since exact day-level matches are too sparse for training. Appendix~\ref{sec:analysis_weak_labels} provides a manual analysis of label quality.

\paragraph{Bi-encoder} First, we use a trained bi-encoder to retrieve event-relevant articles. We adopt CLIP \citep{pmlr-v139-radford21a} as the base model. The bi-encoder retrieves the top-$K$ articles for an image based on the cosine similarity between their embeddings. We use news captions as input to the text encoder. News captions focus on the visual elements directly associated with the event described in the article. They are generated for each article based on its abstract when creating the corpus, as explained in Section \ref{sec:corpus}. We fine-tune the encoders using a symmetric InfoNCE  loss. It computes a softmax over all candidate captions in the batch, encouraging the bi-encoder to produce embeddings with high cosine similarity between the image and the captions of relevant articles.

Each training batch contains pairs of the form (train set image, caption of an event-relevant article). Pairs are added to the batch one at a time. A set of forbidden captions is incrementally updated to ensure that no caption in the batch is event-relevant for more than one image. The bi-encoder should learn representations for all events in the news corpus, not only those covered by the train set images. Hence, we reserve a fraction $n_{\text{random}}$ of each batch for randomly sampled articles from the news corpus that are not event-relevant to any train set image. For these articles, we create a pair using their own image and the corresponding news caption. After each epoch, we evaluate the bi-encoder on the dev set using Recall@K (R@K). We retain the checkpoint corresponding to the best epoch.

\paragraph{Location cross-encoder} 
 The next step is to rerank the articles retrieved by the bi-encoder using a location-focused cross-encoder. For each article, we create a template sentence \textit{``An image from LOCATION''}, where \textit{LOCATION} is a list of the article’s geolocation keywords. We use frozen CLIP image and text encoders. Their embeddings are combined and passed through a trainable linear layer. The output is the similarity score between the image and the article.

Training pairs are created as follows. For each image, we take the top-$K$ articles returned by the bi-encoder. If at least one of them is location-relevant, the image is included in the cross-encoder training data. For each such image, we sample from the top-$K$ one location-relevant article and up to $N_{\text{negative}}$ irrelevant ones, with distinct location keywords, to form training pairs. At the end of each epoch, the cross-encoder is evaluated on the dev set using R@K. We retain the best epoch checkpoint. At inference time, the final location ranking is obtained by multiplying the bi-encoder score by the location cross-encoder score. The pipeline stops here for location prediction.

\paragraph{Article clustering}
The top-$K$ articles retrieved by the bi-encoder are grouped into clusters, based on three rules: (1) all articles within a cluster share at least one location keyword; (2) the temporal span of the cluster, in terms of publication dates, must be less than or equal to $2 \: N_\text{window} + 1$ days; (4) a cluster contains at least $N_\text{min\_size}$ articles. A cluster is relevant for an image if it contains at least one event-relevant article.

\paragraph{Event cross-encoder}
The final step is to rerank the clusters using an event-focused cross-encoder. If the top-$K$ articles retrieved by the bi-encoder yield fewer than two clusters, we skip this step and retain the bi-encoder ranking. For each cluster, we construct a template sentence of the form \textit{``An image between START\_DATE and END\_DATE in LOCATION''}, where \textit{START\_DATE} and \textit{END\_DATE} are the earliest and latest publication dates among the articles in the cluster, and \textit{LOCATION} is the intersection of their geolocation keywords. The model architecture and training procedure are identical to the location cross-encoder. At the end of each epoch, we evaluate the event cross-encoder on the dev set using R@K. We retain the best epoch checkpoint. At inference time, we select the article with the highest bi-encoder score from the highest-ranked cluster, then the article with the highest bi-encoder score from the second-ranked cluster, and so on. Articles that do not belong to any cluster are placed at the end of the new ranking.

\section{Experiments}

\subsection{Datasets}

\paragraph{TARA}
TARA \citep{fu-etal-2022-theres} contains 15,429 news images sourced from \textit{The New York Times} articles. It is divided into train, dev, and test splits. The most recent images are from 2021.

\paragraph{5Pils-OOC}
5Pils-OOC \citep{tonglet-etal-2024-image,tonglet-etal-2025-cove} is a test set containing 624 news images sourced from fact-checking articles. Compared to TARA, 5Pils-OOC includes more images from non-Western regions, particularly India and East Africa, and features 75 images from 2022 and 2023.

\subsection{News article corpus}
\label{sec:corpus}

\begin{figure}
    \centering
    \includegraphics[width=0.9\linewidth]{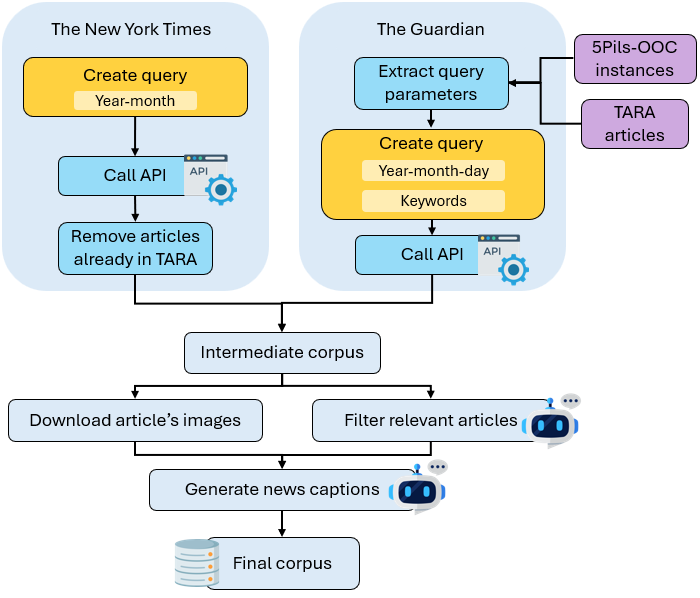}
    \caption{News articles corpus creation process.}
    \label{fig:corpus-creation}
\end{figure}

We collect news articles from two reputable news outlets, \textit{The New York Times} (NY Times) and \textit{The Guardian} (Guardian). The process is illustrated in Figure~\ref{fig:corpus-creation}. Each article is stored as a dictionary containing metadata, headline, lead paragraph, or full text, and image URLs. The corpus is collected using the LLM Qwen2.5-7B \citep{yang2024qwen2}. All prompts are provided in Appendix \ref{sec:prompts}.

\paragraph{NY Times}

We use the official NY Times Archive API,\footnote{\href{https://developer.nytimes.com/docs/archive-product/1/overview}{developer.nytimes.com/docs/archive-product/1/overview}} which returns all articles for a given month. We collect articles from 2010 to 2023, covering the same period as TARA and 5Pils-OOC. Since TARA images originate from NY Times articles, we remove any articles that contain any dev or test images. NY Times articles that contain images from the train set are excluded during training but included for evaluation.

\paragraph{Guardian}

Excluding NY Times articles associated with TARA images during training removes all coverage of certain events.  To compensate for this loss, we query the Guardian API\footnote{\href{https://open-platform.theguardian.com/documentation/}{open-platform.theguardian.com/}} for matching articles. Unlike NY Times, which directly provides all articles for a given month, the Guardian API requires specifying keywords for article search queries. For each TARA image, we use the ground truth date and the keywords from the original NY Times article. Each query returns up to 20 articles. Guardian articles are only used during training.

\paragraph{Filtering}
We retain only articles that describe real-world events that can be reported on with images. We prompt Qwen2.5-7B with eight hand-crafted few-shot examples to classify whether an article describes a visualizable event, based on its news headline. For example, the article \textit{"On nearly every front, President Obama's goal of lower deficits has gotten harder since his first budget a year ago"} is removed from the corpus by the filter. Appendix \ref{sec:manual_analysis_filtering} provides a qualitative analysis of the filtering step.

\paragraph{News caption generation}
The headline of an article often does not provide a clear visual description, and many articles could be illustrated with different types of images. To address this, we generate up to five news captions for each article, describing plausible images that could accompany it. These captions are generated using Qwen2.5-7B, prompted with eight hand-crafted few-shot demonstrations. These captions are used in the bi-encoder to represent the articles.

\paragraph{Corpus statistics and variants} The intermediate corpus contains 415,193 articles. After filtering, the final corpus contains 88,859 articles, of which 77,538 are from the NY Times and 11,321 from the Guardian. For TARA, we exclude all articles published after 2021 because the images do not cover 2022–2023. For 5Pils-OOC, we include all NY Times articles, allowing us to evaluate how well NewsRECON generalizes to images and articles from more recent years without retraining. Appendix \ref{sec:corpus_statistics} provides additional statistics.

\subsection{Metrics}

We use several metrics introduced in prior work on image contextualization.

\textbf{Exact match at K} (EM@K) measures whether at least one of the top-$K$ retrieved items matches the ground truth.

\textbf{Example-F1} (E-F1) \citep{fu-etal-2022-theres} computes a Dice coefficient over the overlap between hierarchical chains derived from the ground truth and the prediction. For instance, the prediction ``Paris, France'', yields the chains \{(Paris, France, Europe), (France, Europe), (Europe)\}.

\textbf{$\Delta$} \citep{tonglet-etal-2024-image} is a date prediction metric whose score is inversely proportional to the absolute distance in years between prediction and ground truth. \textbf{CO$\Delta$} \citep{tonglet-etal-2024-image} is the analogous metric for location, defined as the inverse of the Haversine distance in 1000 km between predicted and ground truth coordinates. The Haversine distance is a standard measure of the great-circle distance between two locations based on their coordinates on a sphere.

\textbf{GREAT} \citep{siingh-etal-2025-getreason} is the weighted average of two similarity scores: one for location and one for date. Both range from 0 to 1, where higher values indicate a prediction closer to the ground truth.

The location score is computed as
\begin{equation*}
\text{GREAT}_{\text{loc}} = \max\left(0, 1 - \frac{d}{1000}\right)
\end{equation*}

where $d$ is the Haversine distance.

The date score at a given granularity $u \in $ \{century, decade, year, month, day\}
is defined as

\begin{equation*}
S_u = 
\begin{cases}
1\text{ if } gt_u = pred_u \text{ else } 0 & \text{if u} = \text{century} \\
\max \left( 0, 1 - \frac{|gt_u - pred_u|}{T_u} \right) &  \text{otherwise}
\end{cases}
\end{equation*}

where $ gt_u $ and $ pred_u $ are the ground truth and predicted values, and $T_u$ is a granularity-specific threshold, e.g., $T_{day}=15$.
The final GREAT$_{\text{date}}$ score is weighted average of the $T_u$ scores.

Unlike \citet{fu-etal-2022-theres}, we do not penalize predictions that are more specific than the ground truth. Instead, we compute the scores starting from the lowest granularity in the ground truth. For example, if the ground truth is a year, we do not consider the day and month predictions. We do this because the lowest ground truth granularity in TARA and 5Pils-OOC is often arbitrary and does not accurately reflect the achievable level of precision.

\begin{table*}[!ht]
\centering
\resizebox{0.87\textwidth}{!}{
  \begin{tabular}{lccccccccccccc}
    \toprule  
       &  \multicolumn{5}{c}{Date} & & \multicolumn{5}{c}{Location} & \\
       \cline{2-6}  \cline{8-12}
      & EM@1 & EM@5 & E-F1 & $\Delta$ & GREAT$_{date}$ & & EM@1 & EM@5 & E-F1 & CO$\Delta$ & GREAT$_{loc}$ & GREAT\\
    \midrule
     CLIP+ (simplified setup) &  (8.1) &  (23.3) & (44.6)  & (38.2) & (55.2) & & (19.4) & (35.0) & (38.7) & (19.7) &  (32.3) & (43.8)  \\
     NewsRECON (simplified setup) &  (\textbf{12.4}) & (\textbf{29.9}) & (\textbf{47.2}) & (\textbf{43.4}) & (\textbf{58.2}) & &  (\textbf{31.5}) & (\textbf{45.1}) & (\textbf{54.9}) & (\textbf{33.0}) & (\textbf{50.0}) & (\textbf{54.1})  \\ 
     \midrule
     \midrule
     NewsRECON &  3.2 & \textbf{9.6} & 42.7 & 41.5 & 58.8 & & 30.0 & \textbf{42.5} & 55.2 & 31.6 & 49.5 & 54.1  \\   
     \midrule
     Text-to-text retriever  & 2.9 & 7.2 & 40.6 & 36.2 & 56.3 & & 11.4 & 21.8 & 38.3 & 12.5 & 30.1 & 43.2  \\
     \citet{DBLP:journals/tomccap/LiuLOSA20} &  0.9 & 4.2 & 37.5 & 32.5 & 54.5 &   & 14.6 & 30.2 & 48.6 &  15.8 & 40.9 & 47.7\\ 
     \midrule
     InternVL3-8B - default &  2.0 & - &  42.0 & 37.5  & 56.3 &  & 25.4 & -  &  48.3 & 26.3  & 41.4 & 48.9  \\
     InternVL3-8B - NewsRECON & 3.4 & - & 44.1 & 41.6 & 59.4 & & 33.9 & - & 58.7 & 36.3 & 55.5 & 57.4 \\
     \midrule
     InternVL3.5-8B - default & 1.5 & - & 38.3  &  32.2 & 52.1 &  & 16.0 & -  &  40.3 & 17.7  & 32.7 &  42.4 \\      
     InternVL3.5-8B - NewsRECON & 3.8 & - & 45.0 & 43.0 & 59.0 & & 33.4 & - & 58.3 & 35.9 & 54.8 & 56.9\\
     \midrule
     Qwen2.5VL-7B - default &  3.2 & - &  39.4 & 40.7 & 59.3  &  &  32.5  & - & 56.7 & 34.2 &  50.5 &54.9 \\
     Qwen2.5VL-7B - NewsRECON & \textbf{4.8} & - & 45.4 & \textbf{45.4} & \textbf{61.3} & & \textbf{35.5} & - & \textbf{60.5} & \textbf{38.4} & \textbf{57.6} & \textbf{59.4}   \\
     \midrule
     Qwen3.5-9B - default &  2.3 & - &  43.4 & 42.0 & 56.8 &  &  26.7  & - &  52.5 & 29.0 & 46.8  & 51.8 \\       
     Qwen3.5-9B - NewsRECON &  4.1 & - & \textbf{46.0} & \textbf{45.4} &  61.0 & & 34.8 & - & 59.7 & 37.5 & 57.0 & 59.0\\ 
     \midrule
     Molmo2-8B - default & 0.8 &  - & 25.7 & 25.2 &  51.0 &  & 16.5 & - & 41.7 & 18.0 & 32.7 & 41.9  \\
     Molmo2-8B - NewsRECON & 3.2 & - & 44.6 & 42.3 & 57.7 &  & 33.3 & - & 57.5 & 35.4 & 54.0 & 55.8\\
  \bottomrule
\end{tabular}
}
  \caption{Main results on the TARA test set (\%).  Results in brackets are obtained under the simplified setup where the choice of labels is limited to the date-location pairs present in the test set.}
  \label{tab:results_tara}
\end{table*}

\subsection{Baselines}

\textbf{CLIP+} \citep{fu-etal-2022-theres} is CLIP model fine-tuned on TARA. It uses a simplified setup in which the candidate date-location pairs are restricted to those present in the test set. To enable a fair comparison, we also report NewsRECON results on TARA after mapping the predicted date to the closest available date in the test set. We do not compare against QR-CLIP~\citep{10.1145/3689638}, which extends CLIP+ by leveraging Wikipedia passages as evidence, as neither the Wikipedia corpus nor the code was publicly available at the time of submission. We implement a \textbf{text-to-text retriever} baseline in which the MLLM Qwen2.5VL-7B \citep{qwen2.5vl} first generates a news caption for the input image; the top-K articles are then retrieved by ranking their caption embeddings against the embedding of the generated caption via cosine similarity, computed using bge-large \citep{bge_embedding}. We also evaluate the cross-modal news retrieval bi-encoder of \citet{DBLP:journals/tomccap/LiuLOSA20}. The text component fuses the embeddings of the article's abstract, caption, and lead paragraph. We use CLIP as the base encoder. The feature fuser and linear transformation layer are trained on TARA, while other layers are frozen.

We also prompt MLLMs to predict the date and location in different settings. We evaluate five popular open-weight MLLMs: InternVL3-8B \citep{internvl3} and Qwen2.5VL-7B \citep{qwen2.5vl}, InternVL3.5-8B \citep{internvl3.5}, Qwen3.5-9B \citep{qwen3.5}, and Molmo2-8B \citep{molmo2}. The \textbf{default} setting is without external evidence. We also study a \textbf{NewsRECON} setting in which the top-3 articles retrieved by NewsRECON serve as evidence to an MLLM. Appendix \ref{sec:scaling} discusses the impact of using smaller and larger models of the InternVL3 and Qwen2.5VL families. Appendix \ref{sec:prompts_sensitivity} discusses two less performant settings from prior work.

On 5Pils-OOC, we further compare NewsRECON to \textbf{COVE} \citep{tonglet-etal-2025-cove}, the SOTA method for image contextualization. COVE leverages RIS evidence to predict the image's date and location. We use the default LLMs of COVE, i.e., LlavaNext-7B \citep{10.5555/3666122.3667638,Liu_2024_CVPR} for object captioning and Llama3-8B \citep{dubey2024llama3} for date and location prediction. It is not meaningful to evaluate COVE on TARA, as the RIS engine would trivially return the NY Times article from which the image was collected.

\subsection{Implementation details}

All models are evaluated in a single run. Appendix \ref{sec:random_seed} evaluates NewsRECON with different seeds.
MLLMs are prompted with a temperature of 0 to make the results more deterministic. All experiments are run on two A100 GPUs with 80GB of memory. Appendix \ref{sec:hyperparameters} provides the hyperparameter values, obtained by tuning on the TARA dev set.

\subsection{Results on TARA}

Table \ref{tab:results_tara} provides the results on the TARA test set.

\textbf{Insight 1: NewsRECON outperforms retrieval baselines.} NewsRECON outperforms CLIP+ under the simplified setup and the text-to-text retriever under the standard evaluation setup. The performance gap between both setups, regardless of the method, suggests that the simplified setup underestimates the task's difficulty. A manual analysis of the text-to-text retriever reveals that most errors stem from overly generic captions generated by Qwen2.5VL-7B, which fail to match specific article headlines; this is despite Qwen2.5VL-7B achieving the strongest date and location prediction performance among all MLLMs.

\textbf{Insight 2: NewsRECON is competitive with MLLM baselines.}
Despite being significantly smaller, NewsRECON outperforms four of the five MLLM baselines in EM@1. Only Qwen2.5VL performs better on the location task by a 2.5 pp margin. The gap in GREAT score of NewsRECON with the MLLMs in the \textit{default} setting ranges from -0.8 pp for Qwen2.5VL to +12.2 pp for Molmo2, with an average of +6.1pp.

\textbf{Insight 3: MLLMs benefit from the articles retrieved by NewsRECON.}
For all MLLMs, the best performance is achieved when retrieved articles are used as external evidence. This leads to large improvements in GREAT score:  from +4.5 pp for Qwen2.5VL to +14.5 pp for InternVL3.5, with an average of +9.7 pp . These results establish the new SOTA and highlight the complementarity of NewsRECON for evidence retrieval and the MLLM for answer generation.

\begin{figure*}[!ht]
    \centering
    \includegraphics[width=\linewidth]{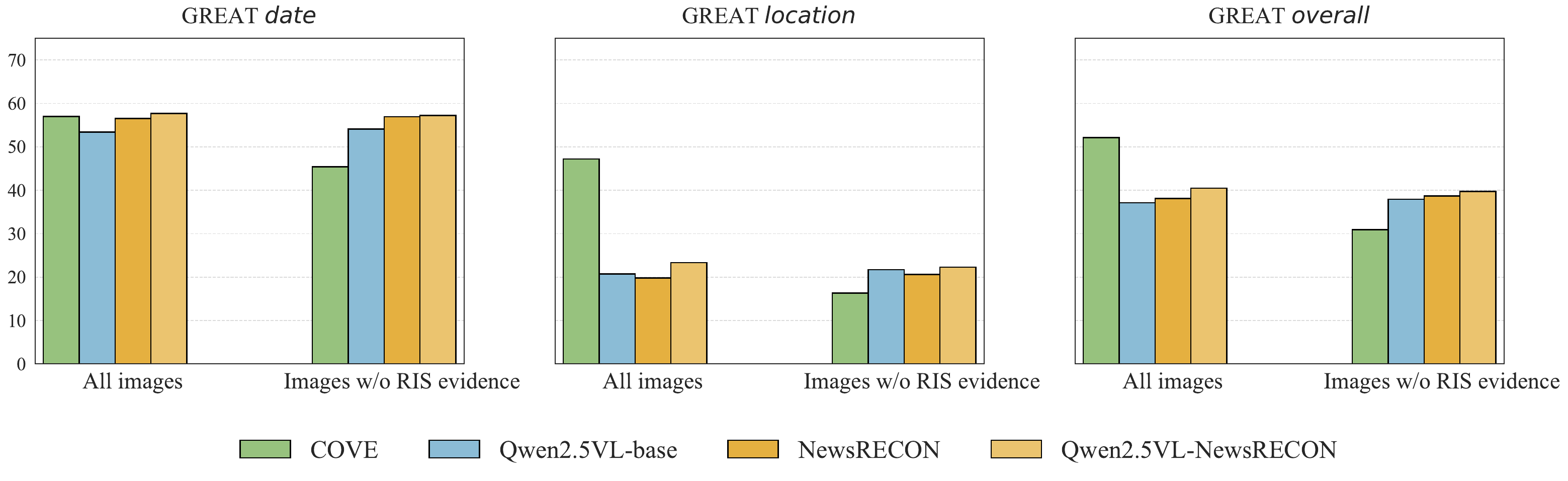}
    \caption{Results on 5Pils-OOC (\%). ``All images'' refers to the entire dataset (N=624), while ``Images w/o RIS evidence'' corresponds to the subset where RIS engines do not return any webpages as evidence (N=172).}
    \label{fig:results_5pils_ooc}
\end{figure*}

\subsection{Results on 5Pils-OOC}

Figure~\ref{fig:results_5pils_ooc} shows the results on 5Pils-OOC using NewsRECON, the best MLLM on the TARA test set, Qwen2.5VL, their combination, and COVE.

\textbf{Insight 4: Qwen2.5VL with NewsRECON outperforms COVE when RIS evidence is unavailable.}
COVE relies on RIS evidence, explaining its overall superior performance. However, NewsRECON was designed for cases where no RIS results are available. In such scenarios, the combination of Qwen2.5VL and NewsRECON outperforms COVE by more than 8 pp in GREAT score, demonstrating that the automated retrieval of relevant news articles is a strong alternative when RIS evidence is missing. This is particularly valuable to human fact-checkers and journalists, as such cases typically require more time and effort to verify.

\textbf{Insight 5: NewsRECON generalizes to out-of-distribution data.}
Despite a performance drop relative to the TARA test set, due to temporal, geographic, and stylistic shifts from the TARA training set, NewsRECON generalizes well to 5Pils-OOC, remaining stronger than Qwen2.5VL for date prediction and slightly weaker for location. Notably, NewsRECON achieves a GREAT score of 41.4 on the subset of 5Pils-OOC images captured in 2022--2023 (n=75), which fall outside the temporal span of the training set, compared to 38.7 on the full 5Pils-OOC test set, demonstrating robustness to events in years unseen during training.

\subsection{Analysis}

The following subsections report our main findings, with additional results provided in Appendix \ref{sec:additional_results}.

\subsubsection{Ablations}

\begin{table}
  \resizebox{\linewidth}{!}{
  \begin{tabular}{lccccc}
    \toprule  
    & \multicolumn{2}{c}{Event (date+location)} & & \multicolumn{2}{c}{Location} \\
    \cline{2-3}  \cline{5-6}
     & R@1 & R@5 & & R@1 & R@5  \\  
    \midrule
   Frozen bi-encoder & 1.4 &  4.0 & & 14.6 & 32.6  \\
   w.  bi-encoder fine-tuning & 1.7 & 4.6 & & 18.6  &  36.0 \\
   w. location cross-encoder & 1.5 & 4.9 & & \textbf{28.6} & \textbf{42.9}  \\
   w. event cross-encoder  & 1.7  & 4.6 & & 16.6 &  32.3 \\   
   w. event cross-encoder + clustering  & \textbf{2.1} & \textbf{5.4} & & 20.2 & 32.7 \\
  \bottomrule
\end{tabular}}
  \caption{R@K of relevant articles on the TARA dev set with different versions of NewsRECON  (\%).}
    \label{tab:ablation}
\end{table}

Table~\ref{tab:ablation} reports ablation results on the TARA dev set using R@K. Fine-tuning the bi-encoder yields a gain of 4 pp in R@1 for location and 0.3 pp for event, highlighting the greater difficulty of retrieving event-relevant articles, which require the correct time window. The location cross-encoder leads to a substantial improvement of 10 pp in R@1 and 6.9 pp in R@5. In contrast, the event cross-encoder with clustering yields smaller gains, 0.4 and 0.8 pp for R@1 and R@5, respectively. Clustering is an important intermediate step for the event cross-encoder. Without it, the performance does not improve over the fine-tuned bi-encoder. For both cross-encoders, the improvement over the fine-tuned bi-encoder exceeds the gains achieved through bi-encoder fine-tuning alone, underscoring their importance in the pipeline. Cross-encoders outperform one another on their respective tasks.

\subsubsection{Impact of input type for bi-encoders}

\begin{table}
\centering
\resizebox{0.9\linewidth}{!}{
  \begin{tabular}{lcc}
    \toprule  
    & Before fine-tuning & After fine-tuning\\
    \midrule
     Abstract & 38.2 &  42.6 \\
     Image & 25.5 &   29.5\\
     News caption & 37.1 &  \textbf{43.7} \\
     
  \bottomrule
\end{tabular}}
    \caption{R@100 of event-relevant articles with different bi-encoder input types on the dev set (\%).}
    \label{tab:input-type}
\end{table}

\begin{figure*}
    \centering
    \includegraphics[width=0.84\textwidth]{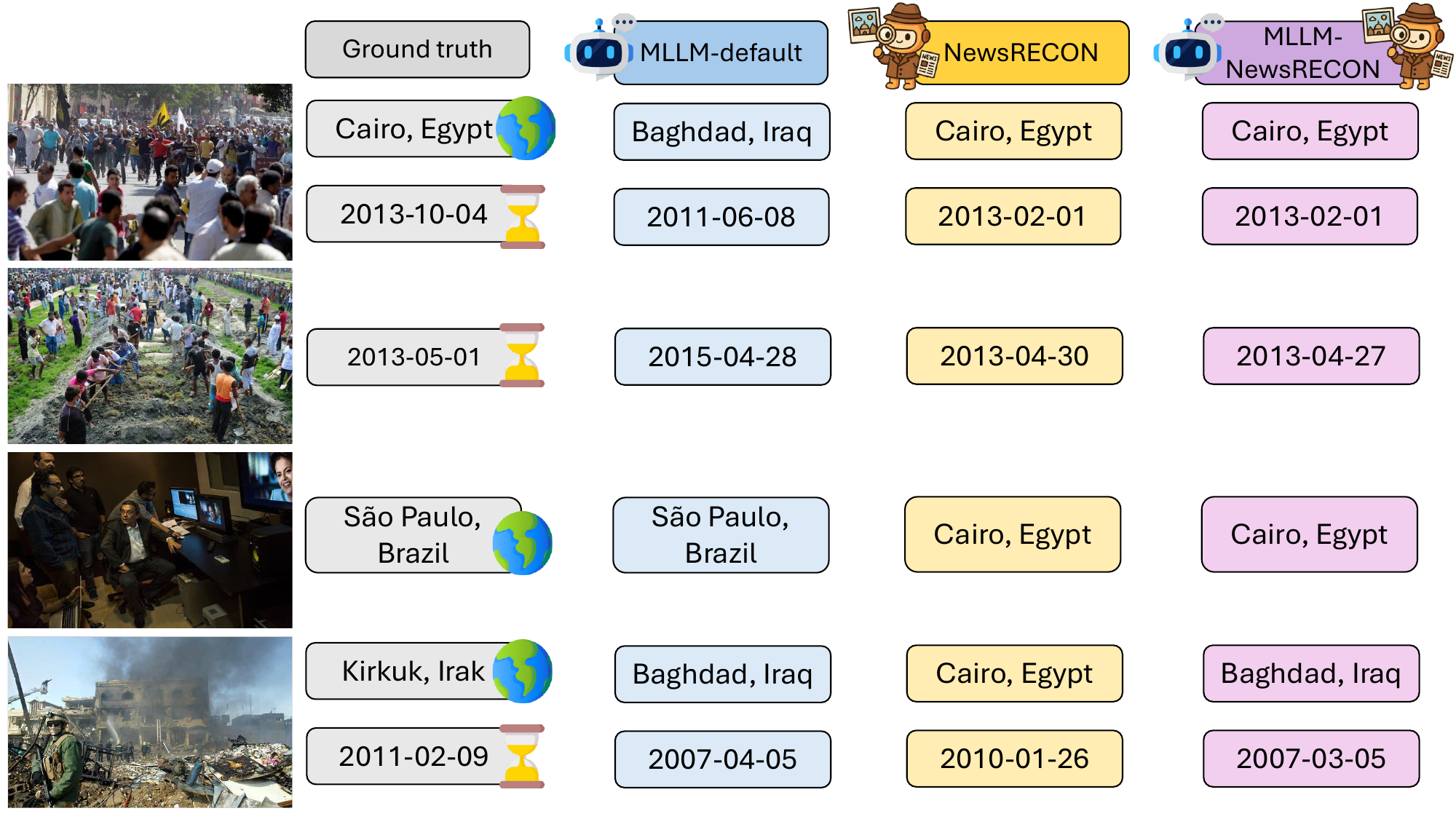}
    \caption{Predictions of different models for four instances of the TARA test set.}
    \label{fig:case-studies}

\end{figure*}

We use news captions to represent each article for the bi-encoder stage. However, other representations are possible, such as the abstract or the image illustrating the article. Table~\ref{tab:input-type} compares the R@100 obtained with different input types on the TARA dev set. While the best input type before fine-tuning is the abstract, the score with news captions is 1.1 pp higher after fine-tuning. We hypothesize that this is because news captions focus on the visual aspects of the event, whereas the abstract is often a more general summary. Using the article's image yields a lower R@100, both before and after fine-tuning. We assume this is because images contain more misleading cues, such as a shared flag between two images, which may correctly indicate the location but not the specific event.

\subsubsection{Impact of corpus size} 

Table~\ref{tab:corpus-size} examines how the size of the retrieval corpus affects GREAT scores. We randomly shuffle the corpus and split it into three. We then evaluate three corpus sizes: one-third, two-thirds, and the full corpus. Reducing the corpus size negatively affects both date and location scores. However, the overall decrease is only 1.2 pp in GREAT score, indicating that the corpus contains redundancy in terms of locations and events, and that a smaller subset is still sufficient to predict many dates and locations accurately. The linear trend for the location score suggests that scaling the corpus beyond its current size could yield small but steady gains in performance. Similarly, the date scores suggest that further improvement can be obtained by scaling the corpus to a more limited extent.

\begin{table}
\centering
  \resizebox{0.9\linewidth}{!}{
  \begin{tabular}{lcccc}
    \toprule  
     &  Size & GREAT$_{date}$ &  GREAT$_{loc}$ & GREAT \\  
    \midrule
    Full  corpus & 77,074 & 59.5 & 51.6 &  55.6 \\
   2/3 of corpus & 51,382 & 58.9 & 50.8 & 54.9 \\
   1/3 of corpus & 25,691 & 58.8 & 50.0 & 54.4 \\  
  \bottomrule
\end{tabular}}
  \caption{GREAT scores on the TARA dev set for different
size of the news articles corpus (\%).}
  \label{tab:corpus-size}
\end{table}

\subsubsection{Qualitative analysis}
\label{sec:error_analysis}

We analyze predictions of Qwen2.5VL, NewsRECON, and their combination, on a random sample of 50 instances from the TARA test set. Figure~\ref{fig:case-studies} presents four examples. We observe 15 incorrect predictions for location and 36 for date. For the location task, all three approaches correctly identify the location in 18 cases. In five cases, the incorporation of NewsRECON’s top-3 retrieved articles enables Qwen2.5VL to revise an initially incorrect prediction into a correct one. However, in six instances, Qwen2.5VL's original correct prediction becomes incorrect due to the retrieved articles. Finally, in three cases, Qwen2.5VL retains its correct prediction even when the retrieved articles are irrelevant. These observations indicate that the retrieved articles should not always be trusted by the MLLM. Nonetheless, the overall quantitative results on the full TARA test set (Table~\ref{tab:results_tara}) demonstrate that combining NewsRECON with Qwen2.5VL yields improvements. For the date task, NewsRECON’s top-3 articles contain the correct date in three cases, and Qwen2.5VL successfully leverages them in two of them. In many cases, while incorrect, the date predicted with the top-3 articles is closer to the ground truth than the default MLLM prediction.

\section{Conclusion}

In this work, we investigated the potential of using news article retrieval to contextualize images in scenarios where RIS engines are unavailable. We evaluated a cross-modal retrieval pipeline, NewsRECON, which queries a corpus of over 85,000 articles, using a fine-tuned bi-encoder and two specialized cross-encoders for location- and event-relevant articles retrieval. The pipeline outperforms prior retrieval baselines on the TARA test set. The top-$K$ retrieved articles can serve as evidence for an MLLM, improving the GREAT score by up to 14.5 pp and achieving the new SOTA. Furthermore, the pipeline generalizes to 5Pils-OOC without retraining, despite geographic and temporal shifts. 

\newpage
\section*{Limitations}
We identify three limitations to our work.

Firstly, the corpus contains articles only from two news outlets. We selected the NY Times and the Guardian because, to the best of our knowledge, they are the only news outlets that provide free and comprehensive APIs for article collection. In future work, including more news outlets from countries beyond the UK and the USA would help expand event coverage. However, this poses some data collection issues as official APIs may not be available.  With a larger corpus, new methodological challenges are likely to arise, such as the need for multilingual retrieval and the risk of introducing more noise.

Secondly, while news article retrieval improves an MLLM's overall performance, it can also alter the MLLM's initial correct predictions due to irrelevant evidence, as shown in Figure \ref{fig:case-studies}. Future work should consider better approaches to combining the MLLM with retrieved news articles, e.g., by directly resolving conflicts among the top 3 pieces of evidence \citep{wang2025retrieval} or between the evidence and information extracted from the image.

Thirdly, the datasets that we used, TARA and 5Pils-OOC, face an important limitation. As new MLLMs get released, their parametric knowledge about past events improves, and these datasets will become increasingly saturated. Future work should address this issue by providing quarterly or annual updates to image contextualization benchmarks.

\section*{Ethics statement}

\paragraph{Intended use} NewsRECON provides a new method for the important problem of contextualizing news images by identifying their date and location. This has several applications, including the early detection of multimodal misinformation on social media. NewsRECON and the accompanying corpus should only be used for non-commercial academic research on image contextualization and multimodal misinformation detection. 

\paragraph{Data access} The datasets and articles were accessed in compliance with the German Act on Copyright and Related Rights (§60d UrhG)
. We release our code under an Apache 2.0 license. TARA and 5Pils-OOC are made available by their authors under Apache 2.0 and CC BY-SA 4.0 licenses, respectively. We do not provide the corpus articles directly. Instead, we provide the code needed to reproduce our corpus collection, along with the URLs of the articles.

\paragraph{AI assistants use}  AI assistants were used in this work to assist with writing by correcting grammar mistakes and typos.

\section*{Acknowledgments}
This work has been funded by the LOEWE initiative (Hesse, Germany) within the emergenCITY center (Grant Number: LOEWE/1/12/519/03/05.001(0016)/72), by the German Federal Ministry of Research, Technology and Space and the Hessian Ministry of Higher Education, Research, Science and the Arts within their joint support of the National Research Center for Applied Cybersecurity ATHENE, and
by the Flanders AI Research Program.
The figures have been designed using resources from Flaticon.com.

\bibliography{bibliography,anthology}

\appendix

\section{NewsRECON pseudocode}
\label{sec:pseudocode}

Algorithm \ref{alg:newsrecon_inference} shows the pseudocode of NewsRECON at inference time.

\begin{algorithm*}[t]
\caption{NewsRECON}
\label{alg:newsrecon_inference}
\begin{algorithmic}[1]

\Require Image $I$, corpus $\mathcal{A}$, bi-encoder $E_{\text{bi}}$, location cross-encoder $C_{\text{loc}}$, event cross-encoder $C_{\text{evt}}$, parameters $K$, $N_{\text{window}}$, $N_{\min\_\text{size}}$.
\\
\Statex \textbf{Bi-encoder Retrieval}
\State  $v_I \gets E_{\text{bi}}^{\text{img}}(I)$
\State  $v_a \gets E_{\text{bi}}^{\text{text}}(a.\text{caption})$ $\forall a \in \mathcal{A}$
\State $s_{\text{bi}}(a) \gets \cos(v_I, v_a)$
\State $\mathcal{A}_K \gets$ sort($s_{\text{bi}}$)[:K]
\\
\Statex \textbf{Location Reranking}
\For{ $a \in \mathcal{A}_K$}
    \State  $t_a \gets $ ``An image from $\{a.\text{locations}\}$''
    \State $s_{\text{loc}}(a) \gets C_{\text{loc}}(I, t_a)$
    \State $s_{\text{comb}}(a) \gets s_{\text{bi}}(a) \cdot s_{\text{loc}}(a)$
\EndFor
\State $\mathcal{R}_{\text{loc}} \gets$ sort($s_{\text{comb}}$)
\\
\Statex \textbf{Clustering of Retrieved Articles}
\State Form clusters $\mathcal{C}$ from $\mathcal{A}_K$ such that:
\State \hspace{0.5cm} (i) articles share $\ge$1 location keyword;
\State \hspace{0.5cm} (ii) temporal span $\le 2N_{\text{window}}{+}1$ days;
\State \hspace{0.5cm} (iii) cluster size $\ge N_{\min\_\text{size}}$.

\If{$|\mathcal{C}| < 2$}
    \State $\mathcal{R}_{\text{evt}} \gets  $ sort($s_{\text{bi}}$)
    \State \Return $\mathcal{R}_{\text{loc}}, \mathcal{R}_{\text{evt}}$
\EndIf
\\
\Statex \textbf{Event Reranking}
\For{ $c \in \mathcal{C}$}
    \State Let $[\text{start}(c), \text{end}(c)]$ be earliest/latest dates in $c$
    \State Let $\text{loc}(c)$ be intersection of location keywords in $c$
    \State $t_c \gets$ ``An image between \{start(c)\} and \{end(c)\} in \{loc(c)\}''
    \State  $s_{\text{evt}}(c) \gets C_{\text{evt}}(I, t_c)$
    \State $a_{\max}(c) \gets \arg\max_{a \in c} s_{\text{bi}}(a)$
\EndFor

\State $\mathcal{C}_{\text{rank}} \gets $ sort($s_{\text{evt}}(c)$)
\State $\mathcal{R}_{\text{evt}} \gets [\, a_{\max}(c) \text{ for } c \in \mathcal{C}_{\text{rank}}\,]$
\State Append unclustered articles in decreasing $s_{\text{bi}}$
\\
\State \Return $\mathcal{R}_{\text{loc}}, \mathcal{R}_{\text{evt}}$

\end{algorithmic}
\end{algorithm*}

\section{Manual analysis of the relevant articles}
\label{sec:analysis_weak_labels}

We manually analyze a sample of 30 instances from the TARA train set in which only one event-relevant article is retrieved. In 90\% of cases, the location and date of the articles match the ground truth location and date of the input image. The three error cases stem from weak location-keyword matching. In 11 out of 30 instances, the dates not only match, but the article also describes the exact same event. Although an exact event match is not required for predicting the correct date and location, it can provide richer contextual information when NewsRECON is combined with an MLLM.

\section{Prompts}
\label{sec:prompts}

Figures \ref{fig:prompt-category} and \ref{fig:prompt-caption} show the prompts provided to the LLM Qwen2.5/7B/ when creating the corpus. 

Figures~\ref{fig:prompt-qa-default-celebrity} and~\ref{fig:prompt-qa-newsrecon} display the prompt templates used for question answering with MLLMs. The final prompt varies depending on the task. For date prediction, we explicitly define the valid date range. The variable MAX\_DATE is set according to the dataset: for TARA, the latest permissible date is December 31, 2021, while for 5Pils-OOC, it is December 31, 2023.

\section{Manual analysis of article filtering}
\label{sec:manual_analysis_filtering}

Qwen2.5-7B is selected for article filtering and captioning due to its strong performance on standard benchmarks \citep{yang2024qwen2}, at the time of corpus collection. We verify the quality of the automated filtering of irrelevant articles with Qwen2.5-7B. We randomly sample 50 articles from the corpus, half of which are predicted as relevant by the LLM, and the other half are predicted as non-relevant. Table \ref{tab:filtering_manual_analysis} provides the results of our human evaluation. We agreed with Qwen2.5-7B's prediction for 82\% of the articles. Although the Qwen2.5-7B's filter does not have perfect accuracy, it enables us to reduce the corpus size by more than 75\%, dramatically decreasing storage and computational costs.

\begin{table}
    \centering
    \begin{tabular}{ccc}
    \toprule
     &  \multicolumn{2}{c}{Qwen2.5-7B label} \\
      \cline{2-3}
       Human label  & Relevant  & Non-relevant \\
         \midrule
     Relevant    & 21  & 5 \\
     Non-relevant &  4 &   20 \\
     \bottomrule
    \end{tabular}
    \caption{Human evaluation of Qwen2.5-7B article filtering on a sample of 50 articles.}
    \label{tab:filtering_manual_analysis}
\end{table}

\section{News articles corpus additional statistics}
\label{sec:corpus_statistics}

Table \ref{tab:corpus-statistics} provides more details about the composition of the corpus for each split of the TARA and 5Pils-OOC datasets.

\begin{table*}
  \resizebox{\textwidth}{!}{
  \begin{tabular}{lcccccc}
    \toprule  
    &  & \multicolumn{3}{c}{NY Times} & &  Guardian \\
     &   & 2010-2021 & TARA train (2010-2021) &  2022-2023 & & 2010-2021   \\  
          \cline{3-5} \cline{7-7} 
     Split & Count & 53,448 &  12,306 & 11,784  &   &  11,321   \\
    \midrule
    TARA train & 64,769 &  \checkmark  &  &   &  &  \checkmark  \\
    TARA dev & 77,075 &  \checkmark  & \checkmark  &  &  &  \checkmark \\
    TARA test & 65,754 &  \checkmark  & \checkmark  &  &    \\
    5Pils-OOC test & 77,538 & \checkmark &  \checkmark &   \checkmark  &      \\
  \bottomrule
\end{tabular}}
  \caption{News articles corpus statistics.}
  \label{tab:corpus-statistics}
\end{table*}

\begin{figure*}
    \centering
    \begin{tcolorbox}[colback=gray!5, colframe=gray!80, title=Relevant article filtering prompt, width=\textwidth]
    You are a helpful assistant for journalism. You are given the headline of a news article which comes with an image. Your task is to classify the news article in one of the two categories based on the content of the headline.\\
    
    Category 1:  the content discusses at least one visual event, i.e., a physical, observable event, and it is very likely that the image accompanying the news article is showing this event.\\
    
    Category 2: the news article is not discussing any visual event, e.g., it discusses only political decisions or the results of the stock exchange, it is an interview, ... it is likely that the image accompanying the article serves the role of a stock image. In other words, the image is likely not at the core of the news article.\\

    \{FEW-SHOT EXAMPLES\} \\

    News article headline: \{HEADLINE\}\\

    Answer only with ``Category 1'' or ``Category 2'':

    \end{tcolorbox}
    \caption{Prompt to filter relevant articles. }
    \label{fig:prompt-category}
\end{figure*}

\begin{figure*}
    \centering
    \begin{tcolorbox}[colback=gray!5, colframe=gray!80, title=News image caption generation prompt, width=\linewidth]
    You are a helpful assistant for journalism. You are given the headline of a news article and your task is to generate 5 news image captions that would be suitable to illustrate this news article. Answer only with the captions as a list of strings.\\

    \{FEW-SHOT EXAMPLES\} \\

    News article headline: \{HEADLINE\}\\

    News image captions:

    \end{tcolorbox}
    \caption{Prompt for news caption generation. }
    \label{fig:prompt-caption}

\end{figure*}

\begin{figure*}
    \centering
    \begin{tcolorbox}[colback=gray!5, colframe=gray!80, title=Question answering prompt (default/celebrities), width=\linewidth]
    You are a helpful assistant for journalism. Your task is to predict the \{DATE/LOCATION\} of the given news image.\\

    \textbf{If celebrities}: The following public figures can be seen in the image \{CELEBRITIES\}.\\

    Leverage the image's content to predict the \{DATE/LOCATION\}.\\

    \textbf{If task is location}: Where was the image taken? Answer only with the city, region, and country, structured as a comma-separted list (city,region,country).\\

    \textbf{If task is date}: When was the image taken? Answer only with a date (yyyy-mm-dd, yyyy--mm, or yyyy), as specific as possible. The date need to be included in the range [1900-01-01, \{MAX\_DATE\}].

    \end{tcolorbox}
    \caption{Prompt for question answering (default and celebrities setting). }
    \label{fig:prompt-qa-default-celebrity}
\end{figure*}

\begin{figure*}
    \centering
    \begin{tcolorbox}[colback=gray!5, colframe=gray!80, title=Question answering prompt (NewsRECON), width=\linewidth]
    You are a helpful assistant for journalism. Your task is to predict the \{DATE/LOCATION\} of the given news image.\\

    Leverage the image's content and relevant additional information to predict the \{DATE/LOCATION\}. The following news articles might be relevant to the events shown in the image. Use them to answer the question in addition to the image's content. They are sorted by order of relevance:\\
    
    \{ARTICLES\}

    \textbf{If task is location}: Where was the image taken? Answer only with the city, region, and country, structured as a comma-separted list (city,region,country).\\

    \textbf{If task is date}: When was the image taken? Answer only with a date (yyyy-mm-dd, yyyy--mm, or yyyy), as specific as possible. The date need to be included in the range [1900-01-01, \{MAX\_DATE\}].

    \end{tcolorbox}
    \caption{Prompt for question answering (NewsRECON setting). }
    \label{fig:prompt-qa-newsrecon}
\end{figure*}

\begin{figure*}
    \centering
    \includegraphics[width=0.8\textwidth]{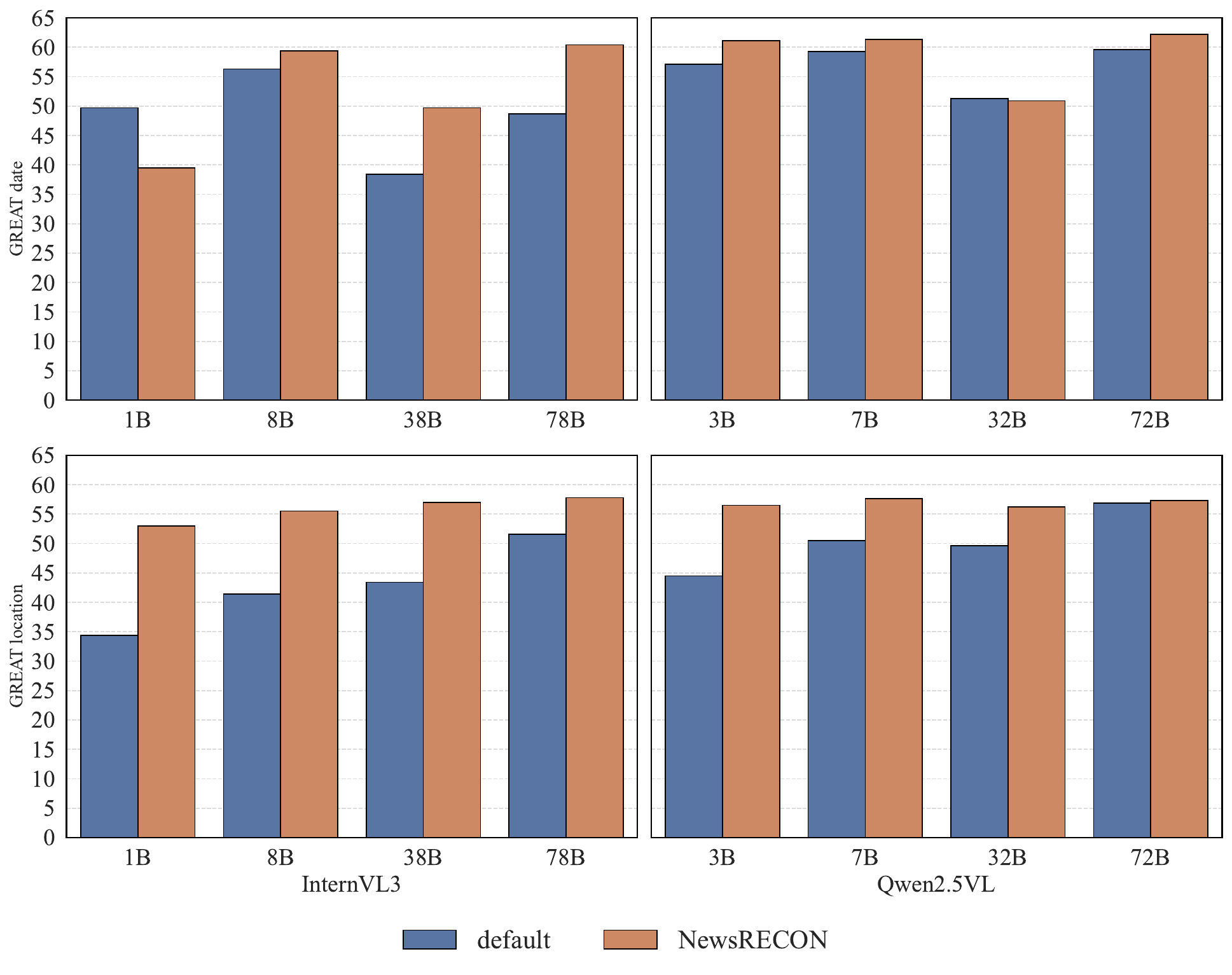}
    \caption{GREAT scores on the TARA test set for different models of the InternVL3 and Qwen2.5VL families using the \textit{default} and \textit{NewsRECON} setting (\%).}
    \label{fig:scaling}
\end{figure*}

\begin{table*}[!ht]
\resizebox{\textwidth}{!}{
  \begin{tabular}{lccccccccccccc}
    \toprule  
       &  \multicolumn{5}{c}{Date} & & \multicolumn{5}{c}{Location} & \\
       \cline{2-6}  \cline{8-12}
      Seed & EM@1 & EM@5 & E-F1 & $\Delta$ & GREAT$_{date}$ & & EM@1 & EM@5 & E-F1 & CO$\Delta$ & GREAT$_{loc}$ & GREAT\\
     \midrule
     123 (default) & 3.2 & 9.6 & 42.7 & 41.5 & 58.8 & & 30.0 & 42.5 & 55.2 & 31.6 & 49.5 & 54.1  \\    
     456 & 3.2  & 9.6  & 42.7 & 41.5  & 58.9 & &  30.0  & 42.5  & 55.2   & 31.6   & 49.5 & 54.2 \\
     789 & 3.2 &  9.6 & 42.7 & 41.5  & 58.8 & &  30.0  & 42.5 & 55.1  & 31.6    & 49.3 & 54.0 \\
  \bottomrule
\end{tabular}
}
  \caption{NewsRECON results on the TARA test set (\%), using three different random seeds. }
  \label{tab:results_seed}
\end{table*}

\section{Impact of MLLM size}
\label{sec:scaling}

Figure \ref{fig:scaling} shows the GREAT scores obtained on the TARA test set by using models of different sizes from the InternVL3 and Qwen2.5VL families in the \textit{default} and \textit{NewsRECON} settings. For almost all model sizes, the \textit{NewsRECON} setting outperforms the \textit{default} setting. The only exception is the GREAT date score for InternVL3-1B. Upon manual inspection of the predictions, we observed that InternVL3-1B in the \textit{NewsRECON} setting often fails to conform to the expected date format, i.e., YYYY, YYYY-MM, or YYYY-MM-DD. Instead, the model provides more verbose explanations of the articles' content. We attribute this failure to comply with instructions given a longer prompt to the smaller size of the MLLM.

Interestingly, performance does not always improve with scale. Qwen2.5-VL has lower scores in its 32B version than in its 7B version, for example. Furthermore, the 72B version only brings limited improvement over the 7B version. This means that image contextualization with MLLMs, whether supplemented with news articles or not, does not benefit from MLLM scaling.

\section{Comparison of different MLLM settings}
\label{sec:prompts_sensitivity}

Table \ref{tab:results_prompts} reports the performance of four different MLLM settings. The scores for Default and NewsRECON settings are the same as in Table \ref{tab:results_tara}. We evaluate two additional settings. The \textbf{celebrity} setting includes in the prompt the list of celebrities detected in the image with the Amazon Rekognition API.\footnote{\href{https://docs.aws.amazon.com/rekognition/latest/dg/celebrities.html}{docs.aws.amazon.com/rekognition/latest/dg/celebrities.html}} 
The \textbf{GETReason} setting consists of multiple intermediate reasoning steps, as originally proposed in~\cite{siingh-etal-2025-getreason}. We apply two modifications to their prompts, resulting in a slight performance improvement. First, we remove the instruction requiring the MLLM to return ``NA'' when uncertain, as we observed that this led the model to overuse ``NA'' even when it could correctly answer the question under the default setting. Second, we revise the prompt for date prediction. While the original prompt requests the model to predict the time period, year, and time of day, it does not explicitly ask for the full date (year, month, and day). We therefore modify the instruction to obtain date-level responses, including day and month.

The results show that adding \textbf{celebrity} information in the prompt improves performance, but not to the extent of the \textbf{NewsRECON} setting. For all models, the \textbf{GETReason} seting is the least performant.

\begin{table*}[!ht]
\centering
\resizebox{0.9\textwidth}{!}{
  \begin{tabular}{lccccccccccccc}
    \toprule  
       &  \multicolumn{5}{c}{Date} & & \multicolumn{5}{c}{Location} & \\
       \cline{2-6}  \cline{8-12}
      & EM@1 & EM@5 & E-F1 & $\Delta$ & GREAT$_{date}$ & & EM@1 & EM@5 & E-F1 & CO$\Delta$ & GREAT$_{loc}$ & GREAT\\
     \midrule
     InternVL3-8B - default &  2.0 & - &  42.0 & 37.5  & 56.3 &  & 25.4 & -  &  48.3 & 26.3  & 41.4 & 48.9  \\
     InternVL3-8B - celebrity & 2.2 & -  &  42.7 & 39.2 &  57.5 &  & 27.9  & - & 50.4 & 28.9 & 44.2 &50.9 \\
     InternVL3-8B - GETReason & 0.6 & - & 27.1  & 15.1  & 33.8 &   &    17.9 & - & 35.1 &  18.9 & 32.1 & 33.0 \\
          InternVL3-8B - NewsRECON & 3.4 & - & 44.1 & 41.6 & 59.4 & & 33.9 & - & 58.7 & 36.3 & 55.5 & 57.4 \\
     \midrule
     InternVL3.5-8B - default & 1.5 & - & 38.3  &  32.2 & 52.1 &  & 16.0 & -  &  40.3 & 17.7  & 32.7 &  42.4 \\     
     InternVL3.5-8B - celebrity & 1.5 & - & 38.9  &  34.0 & 54.5 &  & 19.5 & -  &  40.6 & 20.2  &  26.9 &  40.7 \\  
     InternVL3.5-8B - GETReason & 0.1 & - & 2.5 & 2.1 &  4.2 & & 2.4 & - & 4.3 & 2.6 & 3.8 & 4.0 \\ 
     InternVL3.5-8B - NewsRECON & 3.8 & - & 45.0 & 43.0 & 59.0 & & 33.4 & - & 58.3 & 35.9 & 54.8 & 56.9\\
     \midrule
     Qwen2.5VL-7B - default &  3.2 & - &  39.4 & 40.7 & 59.3  &  &  32.5  & - & 56.7 & 34.2 &  50.5 &54.9 \\
     Qwen2.5VL-7B - celebrity &  3.3 & -  & 40.1 & 41.7  &  59.7 &    & 32.8  & - & 57.0 &  34.5& 51.6 & 55.7 \\
     Qwen2.5VL-7B - GETReason & 2.0 & - &  38.6 &33.6 &  58.5 &  & 32.2 & - & 54.0 & 34.0 &  49.5 & 54.0 \\
     Qwen2.5VL-7B - NewsRECON & \textbf{4.8} & - & 45.4 & \textbf{45.4} & \textbf{61.3} & & \textbf{35.5} & - & \textbf{60.5} & \textbf{38.4} & \textbf{57.6} & \textbf{59.4}   \\
     \midrule
     Qwen3.5-9B - default &  2.3 & - &  43.4 & 42.0 & 56.8 &  &  26.7  & - &  52.5 & 29.0 & 46.8  & 51.8 \\     
     Qwen3.5-9B - celebrity & 2.2 & - & 43.2 & 40.7 & 57.0 & & 28.9 & - & 71.0 &  30.2 & 40.9 & 48.9 \\  
     Qwen3.5-9B - GETReason & 0.9 & - &18.1 & 17.4 & 29.5 & &  19.0 & - & 33.1 & 20.2  & 29.8 & 29.6  \\  
     Qwen3.5-9B - NewsRECON &  4.1 & - & \textbf{46.0} & \textbf{45.4} &  61.0 & & 34.8 & - & 59.7 & 37.5 & 57.0 & 59.0\\  
     \midrule
     Molmo2-8B - default & 0.8 &  - & 25.7 & 25.2 &  51.0 &  & 16.5 & - & 41.7 & 18.0 & 32.7 & 41.9  \\
     Molmo2-8B - celebrity & 0.9 &  - & 30.4 & 29.6 &  51.4 &  & 17.3 & - & 39.2 & 18.6 & 31.5 & 41.5  \\
     Molmo2-8B - GETReason &  0.9 & - & 16.2 & 13.8 & 26.6 & & 15.2  & - & 33.9 & 16.3 & 27.3 & 26.9 \\
     Molmo2-8B - NewsRECON & 3.2 & - & 44.6 & 42.3 & 57.7 &  & 33.3 & - & 57.5 & 35.4 & 54.0 & 55.8\\
  \bottomrule
\end{tabular}
}
  \caption{Main results on the TARA test set for different MLLM settings (\%).}
  \label{tab:results_prompts}
\end{table*}

\section{Impact of random seed}
\label{sec:random_seed}

Table \ref{tab:results_seed} shows the results obtained on the TARA test set using three different random seeds at inference time. NewsRECON achieves similar scores across multiple runs, demonstrating the method's stability.

\section{Implementation details}
\label{sec:hyperparameters}

Table \ref{tab:hyperparameters} provides the hyperparameter values of NewsRECON. These values were obtained after tuning them on the TARA dev set.

\begin{table}[!ht]
  \resizebox{\linewidth}{!}{
  \begin{tabular}{lc}
    \toprule  
    Hyperparameter & Value \\
    \midrule
    \multicolumn{2}{c}{\textit{Bi-encoder}}  \\
    \midrule
    Base model & CLIP/ViT-L-14 \\
    Epochs &  10\\
    Learning rate & 3e-5\\
    Batch size &  256 \\
    $n_{\text{random}}$ & 0.5  \\
    Unfrozen CLIP layers & 4 \\
    Validation metric & R@100 \\
    \midrule
    \multicolumn{2}{c}{\textit{Location cross-encoder}}  \\
    \midrule
    Base model & CLIP/ViT-L-14 \\
    Epochs &  5\\
    Learning rate & 1e-3\\
    Weight decay & 1e-3 \\
    Batch size &  128 \\
    Top-$K$ to rerank & 20  \\
    Validation metric & R@1 \\  
    $N_{\text{negative}}$ & 5 \\
        Embeddings combination & Concatenation\\
    \midrule
    \multicolumn{2}{c}{\textit{Event cross-encoder}} \\
    \midrule
    Base model & CLIP/ViT-L-14 \\
    Epochs &  15\\
    Learning rate & 1e-3\\
    Weight decay & 1e-5 \\
    Batch size &  128 \\
    Top-$K$ to rerank & 50  \\
    Minimum number of clusters & 2 \\
    $N_\text{window}$ & 7  \\
    $N_\text{min\_size}$ & 3 \\
    Validation metric & R@1 \\ 
    $N_{\text{negative}}$ & 5  \\
    Embeddings combination & Concatenation, multiplication, difference \\

  \bottomrule
\end{tabular}

}
  \caption{NewsRECON hyperparameter values}
  \label{tab:hyperparameters}
  
\end{table}

\section{Additional analysis}
\label{sec:additional_results}

\paragraph{Choice of base model} Table \ref{tab:results_encoder} compares the performance of two frozen base models on the TARA dev set, CLIP (\textit{clip-vit-large-patch14}) \citep{pmlr-v139-radford21a} and SIGLIP2 (\textit{siglip2-so400m-patch16-384}) \citep{DBLP:journals/corr/abs-2502-14786}. CLIP's superior performance across all metrics, particularly in location prediction, confirms our decision to use it as the base model for NewsRECON. We hypothesize that the performance gap between the two models is primarily due to the different data on which they were pre-trained. SIGLIP2 has a lower input token limit of 64 tokens compared to CLIP’s 77-token limit. We tokenized all article captions in the corpus to analyze how many were truncated due to SIGLIP2’s token limit. We found that all captions are within the token limit. The mean number of tokens per caption is 27.9 for SIGLIP2 and 29.2 for CLIP. An additional reason to select CLIP is its established strong performance as a base model for related tasks such as out-of-context image detection \citep{DBLP:journals/tcss/PapadopoulosKPP25,10944123}, despite other works showing limited performance on news thumbnail--image alignment \citep{yoon-etal-2024-assessing}.

\begin{table*}[!ht]
\centering
\resizebox{0.85\textwidth}{!}{
  \begin{tabular}{lccccccccccccc}
    \toprule  
       &  \multicolumn{5}{c}{Date} & & \multicolumn{5}{c}{Location} & \\
       \cline{2-6}  \cline{8-12}
      & EM@1 & EM@5 & E-F1 & $\Delta$ & GREAT$_{date}$ & & EM@1 & EM@5 & E-F1 & CO$\Delta$ & GREAT$_{loc}$ & GREAT\\
     \midrule
     SIGLIP2 & 1.0 & 3.4 & 38.3  & 32.0 & 54.3 & & 1.0 & 5.5 & 17.0 & 1.2 & 7.0 & 30.6 \\
     CLIP & 3.9 & 10.5& 43.2 & 41.6 & 59.1 & &  16.4 & 35.4 & 68.3 & 17.3 & 34.2 & 46.6  \\   
  \bottomrule
\end{tabular}
}
  \caption{Performance of CLIP and SIGLIP2 on the TARA dev set (\%).}
  \label{tab:results_encoder}
\end{table*}

\paragraph{Selection of $N_\text{window}$} Table \ref{tab:n_window} shows the performance of NewsRECON for retrieving event-relevant articles, given different values of $N_\text{window}$, which controls the maximum span of a cluster in days. Except when $N_\text{window}$ is 0, corresponding to no clustering, the performance remains similar. We select 3 because it yields a slightly higher R@5 than the other values.

\paragraph{Selection of $N_\text{min\_size}$} Table \ref{tab:n_min_size} shows the performance of NewsRECON for retrieving event-relevant articles, given different values of $N_\text{min\_size}$, which controls the minimum size of a cluster. The value of $N_\text{min\_size}$ has an important effect on R@5. We select a $N_\text{min\_size}$ of 3 as it gives the best R@5.

\begin{table}
\centering
  \resizebox{0.8\linewidth}{!}{
  \begin{tabular}{lccc}
    \toprule  
    & & \multicolumn{2}{c}{Event (date+location)}  \\
    \cline{3-4}  
  $N_\text{window}$ & \# days  & R@1 & R@5  \\  
    \midrule
   0 & 1 &  1.7  & 4.6 \\
   3 & 7 &  \textbf{2.4} & 5.2 \\
   7 (default) & 15 &  2.1 & \textbf{5.4} \\
   14 & 29 &  2.1 & 5.3 \\
  \bottomrule
\end{tabular}}
  \caption{Performance for different values of $N_\text{window}$ on the TARA dev set (\%).}
 \label{tab:n_window}
\end{table}

\begin{table}
\centering
  \resizebox{0.8\linewidth}{!}{
  \begin{tabular}{lcc}
    \toprule  
    &  \multicolumn{2}{c}{Event (date+location)}  \\
    \cline{2-3}  
  $N_\text{min\_size}$ & R@1 & R@5  \\  
    \midrule
   1  & 1.4 & 3.8 \\
   2 &  1.7 & 4.4 \\
   3 (default) &  2.1  & \textbf{5.4} \\
   4  & \textbf{2.2} & 4.7 \\
   5 & 2.1 & 4.9 \\
  \bottomrule
\end{tabular}}
  \caption{Performance for different values of $N_\text{min\_size}$ on the TARA dev set (\%).}
 \label{tab:n_min_size}
\end{table}

\begin{figure*}
    \centering
    \includegraphics[width=\textwidth]{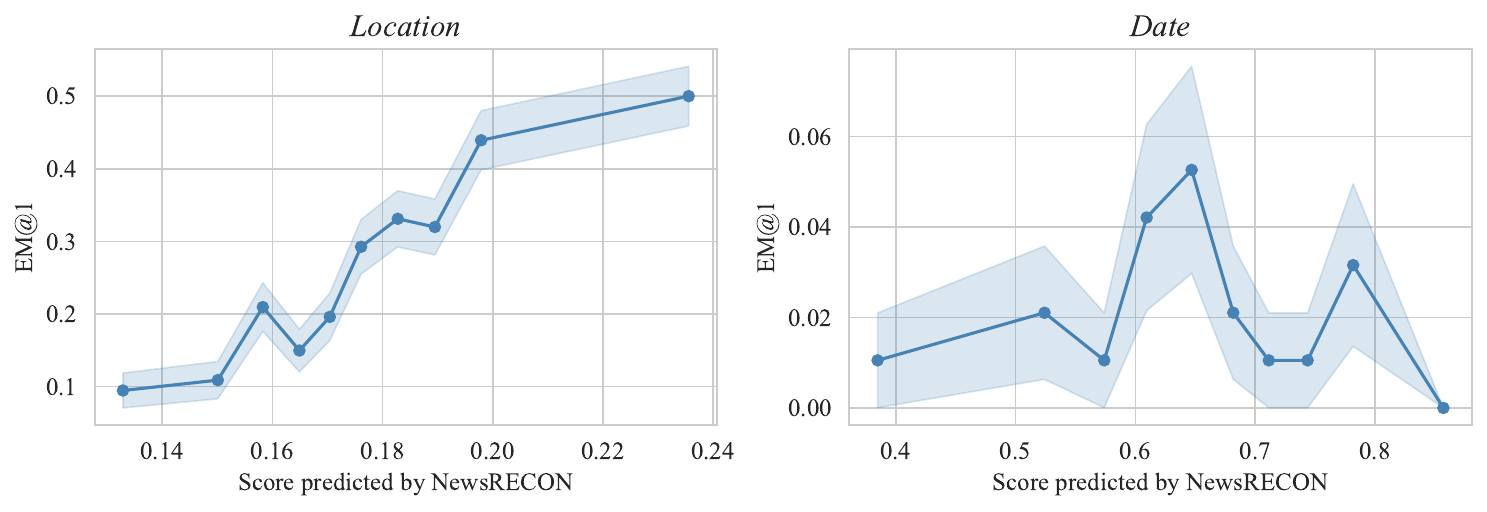}
    \caption{Scores predicted by NewsRECON against the EM@1 on the TARA test set.}
    \label{fig:score-cal}

\end{figure*}

\paragraph{Score calibration} We analyze whether the scores assigned to articles by NewsRECON on the TARA test set are well calibrated. We group the test instances into 10 bins based on the score assigned to the top-1 retrieved article. For each bin, we compute the EM@1. Figure \ref{fig:score-cal} plots the scores predicted by NewsRECON against the EM@1 for both tasks, with one dot per bin. The results show that the scores are well calibrated for the location task: with few exceptions, higher scores translate into higher EM@1. For date prediction, however, the best EM@1 is achieved with scores falling between 0.6 and 0.7, while both higher and lower scores yield worse performance, indicating poorer calibration. This is expected, as date prediction is a more challenging task than location prediction.

\begin{figure*}[!ht]
    \centering
    \includegraphics[width=\textwidth]{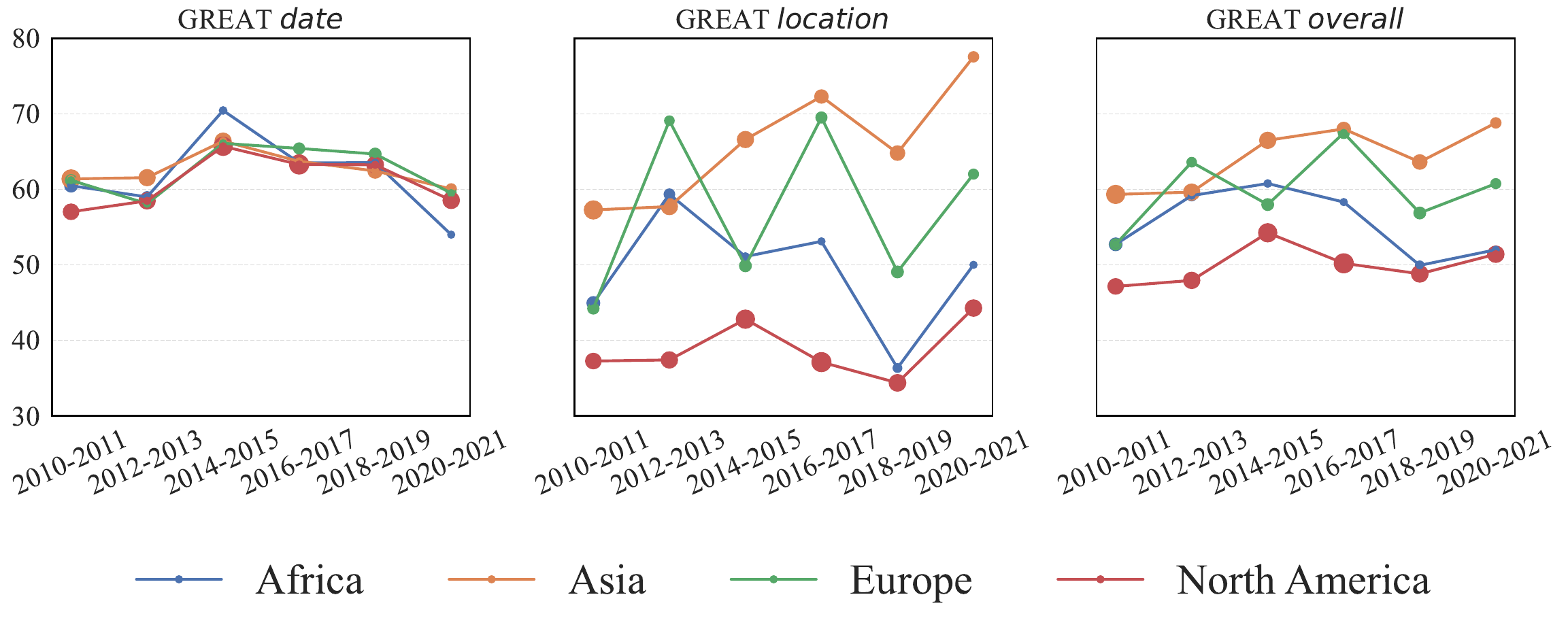}
    \caption{GREAT scores of NewsRECON on the TARA test set by years and continents (\%). The size of the dots is proportional to the number of instances. }
    \label{fig:geo-temp}

\end{figure*}

\paragraph{Time Efficiency} 
At inference time, only the image must be encoded by the bi-encoder, since all article representations are pre-encoded. Encoding an with the bi-encoder takes approximately 0.05 seconds. The location cross-encoder, while significantly improving retrieval performance, introduces an additional average latency of 1.40 seconds, bringing the total inference time to 1.45 seconds. In contrast, the event cross-encoder is computationally efficient because it operates on clusters rather than individual articles. It adds only 0.16 seconds on average, bringing the total to 0.21 seconds to retrieve the top-$K$ event-relevant clusters.

\paragraph{Geographic and Temporal Performance Analysis}
Figure~\ref{fig:geo-temp} presents the GREAT scores of NewsRECON on the TARA test set, broken down by continent and by two-year intervals. The size of each dot is proportional to the number of instances in that bin. The three most represented continents in TARA are North America, Asia, and Europe. We exclude South America and Oceania from the figure due to the very small number of examples, including several bins with zero instances.

For \textit{date prediction}, NewsRECON achieves relatively stable performance across continents and time windows. The highest scores are observed for the 2014–2015 interval.

For \textit{location prediction}, however, performance varies substantially across continents, with GREAT score differences exceeding 30 pp in some cases. The lowest performance is observed for North America. This can be attributed to the granularity of ground truth labels: many images from North America, particularly those from New York City, are annotated with fine-grained locations such as specific neighborhoods. In contrast, many images from Asia and Europe are labeled at broader levels, such as city or country. Since NewsRECON relies on geolocation keywords extracted from article metadata, it is more likely to recover such higher-level locations. Achieving more fine-grained predictions would require a deeper analysis of the article content, for example, by combining NewsRECON with an MLLM. However, for most practical applications of image contextualization, such as debunking multimodal misinformation, a location prediction at the city level is often sufficient in practice \citep{tonglet-etal-2024-image}.

\paragraph{Manual analysis of location and event match} We analyzed two random sample of 30 instances of the TARA test set, one with ground truth location labels and the other with date labels. We manually verified whether the content of the top-3 retrieved articles with NewsRECON matched the image's location or event, respectively. Table \ref{tab:manual_analysis} shows the counts of instances that have one, two, or three matching articles in the top-3 retrieved by NewsRECON. For both tasks, having three out of three matching articles is more common than having one or two. Due to the greater difficulty of the task, event prediction has more cases with zero matching articles than location prediction. Out of the 90 retrieved articles in total, 66.7\% are relevant to the location, and 23.3\% are relevant to the event.

\begin{table}
\centering
  \resizebox{\linewidth}{!}{
  \begin{tabular}{ccc}
    \toprule  
   Number of matching articles & Location & Event (date+location) \\  
    \midrule
    0 & 10 &  23 \\
    1 & 1 & 2 \\
    2 & 7 &  0\\
    3 & 12 & 5 \\
  \bottomrule
\end{tabular}}
  \caption{Number of matching articles in the top-3 retrieved by NewsRECON for a sample of 30 instances of the TARA test set.}
 \label{tab:manual_analysis}
\end{table}

\paragraph{Error analysis case studies} We analyze manually the six cases where a correct prediction by Qwen2.5VL in the \textbf{base} setting is replaced by an incorrect one in the \textbf{NewsRECON} setting. In three cases, the MLLM can, by default, identify the location based on a celebrity or a text written on a building. These are small visual details that are harder to capture with NewsRECON. After changing the setting, the MLLM defaults to the locations corresponding to the retrieved articles. In one case, NewsRECON pairs the image with articles about China because it shows an East Asian emergency doctor. However, all other people in the image are from the Caribbean, suggesting it shows Haiti’s earthquake. On the other hand, the MLLM was able to analyze the full scene by default and provide the correct location. In two other cases, the retrieved articles are relevant, but the MLLM ignores them and makes an incorrect prediction.

\end{document}